\documentclass{article}

\usepackage{microtype}
\usepackage{graphicx}
\usepackage{subfigure}
\usepackage{booktabs} 

\usepackage{subfiles}

\usepackage{amsmath,amsfonts,bm}
\usepackage{mathabx}

\def\eqref#1{equation~\ref{#1}}

\def\1{\bm{1}}

\DeclareMathAlphabet{\mathsfit}{\encodingdefault}{\sfdefault}{m}{sl}
\SetMathAlphabet{\mathsfit}{bold}{\encodingdefault}{\sfdefault}{bx}{n}

\newcommand{\R}{\mathbb{R}}

\newcommand{\softmax}{\mathrm{softmax}}
\newcommand{\Att}{\mathrm{Att}}
\newcommand{\MultAtt}{\mathrm{MultAtt}}

\newcommand{\jac}{\mathbf{D}}

\def\XX{\mathbb{X}}
\def\YY{\mathbb{Y}}

\usepackage{multirow}
\usepackage{hyperref}
\usepackage{url}
\usepackage{graphicx,wrapfig} 
\usepackage{amsmath}
\usepackage{amsthm}
\usepackage{nccmath}
\usepackage{bbm}
\usepackage{xspace}

\newcommand{\MN}{LipschitzNorm\xspace}

\newcommand{\BEAS}{\begin{eqnarray*}}
\newcommand{\EEAS}{\end{eqnarray*}}
\newcommand{\BEA}{\begin{eqnarray}}
\newcommand{\EEA}{\end{eqnarray}}
\newcommand{\BEQ}{\begin{equation}}
\newcommand{\EEQ}{\end{equation}}
\newcommand{\BIT}{\begin{itemize}}
\newcommand{\EIT}{\end{itemize}}
\newcommand{\BNUM}{\begin{enumerate}}
\newcommand{\ENUM}{\end{enumerate}}
\newcommand{\BA}{\begin{array}}
\newcommand{\EA}{\end{array}}

\newcommand{\RR}{\mathbb{R}}

\newcommand{\Eq}[1]{Eq.~(\ref{eq:#1})}
\newcommand{\Sec}[1]{Sec.~\ref{sec:#1}}
\newcommand{\Fig}[1]{Fig.~\ref{fig:#1}}
\newcommand{\Theorem}[1]{Theorem~\ref{th:#1}}
\newcommand{\Lemma}[1]{Lemma~\ref{lemma:#1}}

\newtheorem{lemma}{Lemma} 
\newtheorem{theorem}{Theorem} 
\newtheorem{corollary}{Corollary} 
\newtheorem{remark}{Remark} 

\usepackage{hyperref}

\usepackage[accepted]{icml2021}

\icmltitlerunning{Lipschitz Normalization for Self-Attention Layers with Application to Graph Neural Networks}

\begin{document}

\twocolumn[
\icmltitle{Lipschitz Normalization for Self-Attention Layers \\with Application to Graph Neural Networks}



\icmlsetsymbol{equal}{*}

\begin{icmlauthorlist}

\icmlauthor{George Dasoulas}{huawei,x}
\icmlauthor{Kevin Scaman}{huawei}
\icmlauthor{Aladin Virmaux}{huawei}

\end{icmlauthorlist}

\icmlaffiliation{x}{DaSciM, LIX, École Polytechnique, France}
\icmlaffiliation{huawei}{Noah’s Ark Lab, Huawei Technologies France}

\icmlcorrespondingauthor{George Dasoulas}{george.dasoulas1@gmail.com}

\icmlkeywords{Machine Learning, ICML}

\vskip 0.3in
]



\printAffiliationsAndNotice{}  

\begin{abstract}

Attention based neural networks are state of the art in a large range of applications. However, their performance tends to degrade when the number of layers increases.
In this work, we show that enforcing Lipschitz continuity by normalizing the attention scores can significantly improve the performance of deep attention models.
First, we show that, for deep graph attention networks (GAT), gradient explosion appears during training, leading to poor performance of gradient-based training algorithms.
To address this issue, we derive a theoretical analysis of the Lipschitz continuity of attention modules 
and introduce \MN{}, a simple and parameter-free normalization for self-attention mechanisms that enforces the model to be Lipschitz continuous.
We then apply \MN{} to GAT and Graph Transformers and show that their performance is substantially improved in the deep setting (10 to 30 layers).
More specifically, we show that a deep GAT model with \MN{} achieves state of the art results for node label prediction tasks that exhibit long-range dependencies, while showing consistent improvements over their unnormalized counterparts in benchmark node classification tasks.

\end{abstract}

\section{Introduction}

Over the last few years, attention models became extremely popular in a wide variety of deep learning applications. These architectures made their first appearance in natural language processing and neural machine translation~\citep{bahdanau2014neural, gehring2017convolutional,transformer}, and gradually became state-of-the-art in multiple machine learning tasks, including sequential data learning~\cite{radford, luong, yang}, graph classification~\cite{gat,li2016gated} and computer vision~\cite{xuattend}.
%
Notably, \citet{transformer} showed that efficient deep learning models could be created using attention layers only, leading to the \emph{Transformer} architecture.
Compared to convolutional or linear layers, attention layers have the advantage of allowing the selection of key features in the data while being amenable to backpropagation and gradient descent schemes.

Unfortunately, attention models tend to suffer from poor performance when their depth increases, and most applications have a relatively small number of layers (e.g. $6$ for the Transformers in \citealp{transformer}). While depth is not necessarily synonymous with increased performance, deep architectures showed extremely good performance in many difficult tasks (e.g. image classification) that exhibit complex structural information~\cite{verydeepcnns}. Although there are cases of NLP models, such as GPT-3~\citep{gpt3}, that can scale to very deep architectures (up to 96 layers), for graph attention models state-of-the-art architectures remain shallow and building deep architectures remains an open problem. In the case of graph neural networks (GNNs)~\citep{hamiltonsurvey}, the model depth is directly related to the neighborhood size on which the model aggregates information. In such a case, shallow neural networks are fundamentally unable to capture long-range characteristics, and designing deep graph neural networks is thus a subject of extensive research~\citep{li2018deeper,deepgcns,deepergcn,loukas2020}.

In this work, we show that enforcing Lipschitz continuity by normalizing the attention scores can significantly improve the performance of deep attention models.
To do so, we present \emph{\MN{}}, a normalization scheme for self-attention layers that enforces Lipschitz continuity, and apply this normalization to attention-based GNNs, including graph attention networks (GAT) \citep{gat}, and graph transformers (GT) \citep{NEURIPS2019_9d63484a, unimp}.
Moreover, we show that, without normalization, gradient explosion appears in these architectures due to a lack of Lipschitz continuity of the original attention mechanism~\cite{kim2020lipschitz}.
Finally, we show that such a normalization allows to build deeper graph neural networks that show good performance for node label prediction tasks that exhibit long-range dependencies. The source code is publicly available on Github: \href{https://github.com/gdasoulas/LipschitzNorm}{https://github.com/gdasoulas/LipschitzNorm}.

The remainder of the paper is structured as follows: in \Sec{related}, we provide an overview of the related work on attention mechanisms and graph learning models. Then, in \Sec{definitions}, we provide precise definitions for Lipschitz continuity and attention models. In \Sec{lipschitz.normalization}, we present our theoretical analysis and in \Sec{lipnorm} we introduce our normalization layer, called \emph{\MN{}}. Then, we empirically show in \Sec{explosion} the connection between Lipschitz continuity and gradient explosion during training. These are followed by the experimental evaluation in \Sec{exps}.

\section{Related Work}\label{sec:related}

Initially designed to extend the capabilities of recurrent neural networks~\citep{bahdanau2014neural}, attention models rapidly became a highly efficient and versatile model for machine learning tasks in natural language processing~\citep{radford, yang}, computer vision~\citep{xuattend} and recommender systems~\citep{yingattend}. Recently, novel attention models have been introduced in graph-based systems showing state-of-the-art performance on graph classification~\citep{jbleegraphclassification}, node classification~\cite{gat,unimp} and link prediction~\citep{zhaoattend} tasks. 

\paragraph{Attention and Lipschitz Continuity:}
Although the attention models gain more attraction, little progress has been made in the theoretical study of the attention. ~\citet{perez2018on} showed how attention-based models can be Turing complete and ~\citet{Cordonnier2020On} studied the relationship of self-attention layers and the convolutional networks for image processing.  One important direction that can help towards the expressivity of attention models is the analysis of Lipschitz continuity. Even though the computation of tight Lipschitz bounds of neural networks has been proven to be a hard task~\citep{autolip}, a few approaches suggested Lipschitz-based normalization methods for neural networks~\citep{miyato2018spectral, lip_regularization}. \citet{kim2020lipschitz} showed that the standard dot-product self-attention is not Lipschitz continuous, proposing an alternative attention layer that satisfies the Lipschitz continuity. The latter work assumes that the input and output dimensions of the transformer are equal. Such an assumption is only applicable to Transformer-based models, while for example in graph attention it does not hold, since the inputs are taken neighbor-wise and, thus, with variable length.


\paragraph{Attention and Graph Neural Networks:} In this work, we study Lipschitz properties of the general form of self-attention from the optimization perspective. We propose a normalization that enforces the attention layer to be Lipschitz and prevents the model from gradient explosion phenomena. Graph Neural Networks (GNNs) is a class of models that suffer from gradient explosion and vanishing as the model depth increases and, thus enforcing the Lipschitz continuity of deep attention-based GNNs can enhance their expressivity. Due to the recent success of GNNs in various real-world applications, there is a growing interest in their expressive power, either investigating how GNNs can be universal approximators~\citep{xupowerful, clip2019, maron2018} or studying the impact of the depth and width of the models~\citep{li2018deeper, loukas2020}. The second aspect of the depth analysis still has a few unanswered questions, as the majority of the current state-of-the-art models employ shallow GNNs.  
\paragraph{Depth in GNNs:}
\citet{pairnorm} related the expressivity of graph convolutional networks with the \textit{laplacian oversmoothing} effect and proposed a normalization layer as a way to alleviate it.  More recently, \citet{dropedge} proposed an edge dropping framework on node classification tasks, in order to tackle over-fitting and over-smoothing phenomena and have shown empirically a constant improvement on the original datasets. \citet{deepgcns} and \citet{deepergcn}
introduced frameworks of adaptive residual connections and generalized message-passing aggregators that allow for the training of very deep GCNs. Finally, \citet{loukas2020} studied the effect of the depth and the width of a graph neural network model and  \citet{alon2020} introduced the  \textit{oversquashing} phenomenon as a deterioration factor to the performance of the GNNs. However, to our knowledge, no study on the explicit relationship between the gradient explosion and the GNNs has yet been made.

\section{Notations and Definitions}
\label{sec:definitions}
In this section, we recall the definitions of attention models as well as Lipschitz continuity. This notion will be central in our analysis and help us understand why gradient explosion appears when training attention models (see \Sec{explosion}).

\subsection{Basic Notations}
For any matrix $M\in\R^{n\times m}$, we will denote as spectral norm $\|M\|_*$ its largest singular value, 
$(\infty,2)$-norm $\|M\|_{(\infty,2)} = \max_i (\sum_j M_{ij}^2)^{1/2}$, and Frobenius norm $\|M\|_F=(\sum_{i,j}M_{ij}^2)^{1/2}$.
Moreover, for $\XX$ (resp. $\YY$) a vector space equipped with the norm $\|\cdot\|_\XX$ (resp. $\|\cdot\|_\YY$), the operator norm of a linear operator $f:\XX\to\YY$ will denote the quantity $\vvvert f\vvvert_{\XX,\YY} = \max_{x\in X} \|f(x)\|_\YY / \|x\|_\XX$ and $\vvvert f\vvvert_\XX = \vvvert f\vvvert_{\XX,\XX}$.
Finally, the (Fr\'echet) derivative of a function $f:\XX\to\YY$ at $x\in\XX$ will denote (when such a function exists) the linear function $\jac f_x:\XX\to\YY$ such that, $\forall h\in\XX$, $f(x+h)-f(x) = \jac f_x(h) + o(\|h\|)$.

\subsection{Lipschitz Continuity}
A function $f:\XX \to \YY$ is said to be Lipschitz continuous if there exists a constant $L$ such that, for any $x, y \in \XX$, $\|f(x) - f(y)\|_\XX \leq L \|x - y\|_\YY$. The Lipschitz constant $L_{\XX,\YY}(f)$ will denote the smallest of such constants. Moreover, a Lipschitz continuous function $f$ is derivable almost everywhere and (see \citet[Thm 3.1.6]{alma991000471099706336})
\BEQ
L_{\XX,\YY}(f) = \sup_{X\in\XX} \vvvert\jac f_X\vvvert_{\XX,\YY}\,.
\EEQ

The Lipschitz constant controls the perturbation of the output given a bounded input perturbation, and is a direct extension of the gradient norm to the multi-dimensional case. Indeed, when $f$ is scalar-valued and differentiable, we have $\jac f_x(h) = \nabla f(x)^\top h$ and $\vvvert\jac f_x\vvvert_F = \|\nabla f(x)\|_2$.
In our analysis, we will only consider the Lipschitz constant of attention layers for the Frobenius norm (i.e. the $L_2$-norm of the \emph{flattened} input and output matrices), and derive upper bounds from the previous formula 
(see \Sec{lipschitz.normalization}).


\subsection{Attention Models}\label{sec:attentiondef}
An \emph{attention layer} is a soft selection procedure that uses scores to choose which input vectors to focus on. Before presenting attention layers in their most general form, we first focus on the more simple case with a single vector output in order to provide more intuition to the reader. 

\paragraph{Single Output Case:} Let $x_1, \dots, x_n\in\RR^d$ be a set of input vectors, and $g:\R^d\to\R$ a score function. Each vector is assigned a score $g(x_i)$ that measures the impact of the input vector on the output through a \emph{softmax} function:
\BEQ
\Att(x) = \sum_{i=1}^n \frac{e^{g(x_i)}}{\sum_{j=1}^n e^{g(x_j)}} x_i\,.
\EEQ
In most applications, the score function is linear $g(x)=q^\top x$ where $q\in\R^d$ is a query vector that indicates the direction favored by the attention model.

\paragraph{General Case:}
In many applications, the output is not a single vector, but a collection of vectors. We thus switch to a matrix notation in order to simplify the definitions.
Let $X \in \R^{d\times n}$ be an input matrix whose rows are the input vectors $x_1,\dots,x_n\in\R^d$.
A score function $g:\R^{d\times n}\to\R^{m\times n}$ takes the input matrix and returns scores for each output vector $i\in\{1,\dots,m\}$ and each input vector $j\in\{1,\dots,n\}$. This score is usually linear or quadratic ; however, we will see in \Sec{lipschitz.normalization} that such a generalisation allows to consider more advanced score functions, including overall normalization by a scalar.
The probability weights are then computed using a (row-wise) softmax operator $\softmax: \RR^{m\times n}\to\RR^{n\times m}$ taking as input a score matrix $M\in\RR^{m\times n}$,
\BEQ \label{eq:softmax}
\softmax(M)_{ij} = \frac{e^{M_{ij}}}{\sum_{k=1}^n e^{M_{ik}}}\,.
\EEQ
Note that all rows sum to one, and all coordinates are between $0$ and $1$. Each row can thus be interpreted as a probability distribution over the $n$ input vectors.
Finally, the overall attention module $\Att:\R^{d\times n}\to\R^{d\times m}$ returns a matrix whose columns are weighted averages of the inputs:
\BEQ \label{eq:attention}
\Att(X) = X\, \softmax(g(X))^\top\,.
\EEQ



\paragraph{Multi-Head Attention:}
In order to augment the power of attention models, a common trick consists in concatenating multiple independent attention models. These \emph{multi-head} models can thus focus on multiple directions of the input space at the same time, and are generally more powerful in practice.
A standard procedure consists in first projecting the input vectors into multiple low-dimensional spaces, and combining the results of all attention layers using a linear function.
Let $d_I$ (resp. $d_O$) be the input (resp. output) dimension, $h$ the number of heads, and $W_1, \dots, W_h\in\RR^{d_I\times d}$ and $W_O\in\RR^{d_O\times dh}$ be $h+1$ matrices, then
\BEQ\label{eq:multiheadatt}
\MultAtt(X) = W_O \bigg(\Att(W_1 X)\bigg|\bigg|\dots\bigg|\bigg|\Att(W_h X)\bigg)\,,
\EEQ
where the $||$ operator denotes row-wise concatenation.
In this work, we will consider each attention head separately, using the fact that the Lipschitz constant of multi-head attention can be bounded by that of each attention head.
\begin{theorem}\label{th:multiheadatt}
If each attention head is Lipschitz continuous, then multi-head attention as defined in \Eq{multiheadatt} is Lipschitz continuous and
\BEQ
L_F(\MultAtt) \leq L_F(\Att) \|W_O\|_* \sqrt{\sum_{k=1}^h \|W_k\|_*^2}\,.
\EEQ
\end{theorem}

\paragraph{Transformer Case:}
For Transformers, $m=n$ and the input matrix is decomposed as $X=(Q||K||V)$, where $Q,K,V\in\RR^{d\times n}$ represent, respectively, \emph{queries}, \emph{keys} and \emph{values}. The attention model is then
\BEQ
\Att(X) = V \softmax\left(\frac{Q^\top K}{\sqrt{d}}\right)^\top\,.
\EEQ
Note that the softmax is not multiplied by the whole input vector $X$, but only the values $V$. This is equivalent to projecting the output vectors on a subspace, and thus does not lead to an increase in the Lipschitz constant.

\section{The Lipschitz Constant of Attention}
\label{sec:lipschitz.normalization}

As their name suggest, the purpose of attention layers is to select a small number of input vectors (softmax probabilities tend to focus most of their mass on the largest score). Unfortunately, large scores also tend to create large gradients. In order to show this behavior, we first provide a computation of the norm of the derivative of attention layers. 
Note that the proofs of all lemmas and theorems are provided in the supplementary material.

\paragraph{Derivative of Attention Models:} A direct computation using the definition of \Eq{attention} and the chain rule gives
\begin{multline}
\label{eq:derivative}
\jac \Att_X(H) = H \softmax(g(X))^\top\\
+ X \jac\softmax_{g(X)}(\jac g_X(H))^\top\,,
\end{multline}
where $H\in\RR^{d\times n}$ is an input perturbation. We handle both terms separately, leading to the following upper bound on the Lipschitz constant.
\begin{lemma}\label{lemma:derivative_bound}
For any $X\in\R^{d\times n}$, the norm of the derivative of attention models (see \Eq{attention}) is upper bounded by:
\begin{multline}\label{eq:lf_bound1}
\vvvert\jac \Att_X\vvvert_F \leq \|\softmax(g(X))\|_F \\ + \sqrt{2}\|X^\top\|_{(\infty,2)}\vvvert\jac g_X\vvvert_{F, (2,\infty)}\,.
\end{multline}
\end{lemma}
\Eq{lf_bound1} shows that the Lipschitz constant is controlled by two terms: the first one is related to the \emph{uniformity} of the softmax probabilities, while the second one is related to the size of the input and gradient of the score function. In what follows, we will examine these two terms and show that normalizing the scores by a well-chosen scalar allows to control both simultaneously.

\paragraph{Uniformity of the Softmax Probabilities:}
The first term in \Eq{lf_bound1} is directly related to \emph{how far} the softmax probabilities are from being uniform. More precisely, we have
\BEQ
\|\softmax(g(X))\|_F = \sqrt{\frac{m + \sum_{i=1}^m d_{\chi^2}(S_i, U_n)}{n}}\,,
\EEQ
where $S_i$ is the $i$-th row of $\softmax(g(X))$, $U_n$ is the uniform distribution over $n$ elements, and $d_{\chi^2}(p,q)=\sum_i q_i(p_i/q_i - 1)^2$ is the $\chi^2$-divergence between $p$ and $q$ \citep{fdivergences}. Hence, if all attention heads have uniform probabilities, then $d_{\chi^2}(S_i, U_n)=0$ and $\|\softmax(g(X))\|_F=\sqrt{m/n}$. On the contrary, the distances are maximum when the whole mass of the probabilities is on one element, and in such a case $d_{\chi^2}(S_i, U_n)=n-1$ and $\|\softmax(g(X))\|_F=\sqrt{m}$.
\begin{lemma}
\label{lemma:softmax.frobenius.bounds}
For any $M \in \RR^{m \times n}$, we have
\BEQ
\sqrt{m/n} \leq \|\softmax(M)\|_F \leq \sqrt{m}\,.
\EEQ
\end{lemma}
When $m\gg 1$ (e.g. $m=n$ for Transformers), this implies that the gradients of attention models can be large and lead to the explosive phenomena observed in \Sec{explosion}.
Fortunately, controlling the scale of the scores is sufficient to control the uniformity of the probabilities.
\begin{lemma}
If all the scores are bounded by $\alpha\geq 0$, i.e. for all $i\in\{1,\dots,m\}$ and $j\in\{1,\dots,n\}$, $|g(x)_{ij}|\leq \alpha$, then
\BEQ
\|\softmax(g(X))\|_F \leq e^{\alpha}\sqrt{\frac{m}{n}}\,.
\EEQ
\end{lemma}

Hence, the first objective of the normalization is to scale the scores in order to avoid softmax probabilities to put their entire mass on a single vector.

\paragraph{Impact of a Scalar Normalization:}
Without any additional control, \Lemma{derivative_bound} does not prove the Lipschitz continuity of attention models, as the second term is proportional to the norm of the input matrix $\|X^\top\|_{(\infty,2)}$. For example, if the scores are linear, then their derivative is constant, and the second term in \Eq{lf_bound1} is not bounded.
In order to address this issue, we propose to normalize the score function by a scalar function $c:\RR^{d\times n}\to\RR_+$:
\BEQ\label{eq:Cnorm}
g(X) = \frac{\tilde{g}(X)}{c(X)}\,,
\EEQ
where $\tilde{g}$ is the original score function, and $g$ the normalized one. When $c(X)$ is chosen wisely, this simple normalization is sufficient to obtain a tight bound on the Lipschitz constant of the attention (see \Sec{lipnorm}).

\begin{theorem}\label{th:general_case}
Let $\alpha\geq 0$. If, for all $X\in\RR^{d\times n}$, we have
\BNUM
\itemsep0em
\item[(1)] $\|\tilde{g}(X)\|_{\infty}\leq \alpha c(X)$,
\item[(2)] $\|X^\top\|_{(\infty,2)}\vvvert\jac \tilde{g}_X\vvvert_{F,(2,\infty)}\leq \alpha c(X)$,
\item[(3)] $\|X^\top\|_{(\infty,2)}\vvvert\jac c_X\vvvert_{F,1}\|\tilde{g}(X)\|_{(2,\infty)}\leq \alpha c(X)^2$,
\ENUM
then attention models (see \Eq{attention}) with score function $g(X) = \tilde{g}(X) / c(X)$ are Lipschitz continuous and
\BEQ
L_F(\Att) \leq e^{\alpha}\sqrt{\frac{m}{n}} +  \alpha\sqrt{8}\,.
\EEQ

\end{theorem}

First, note that $\alpha$ controls the scale of all the scores, as assumption (1) implies $\|g(X)\|_\infty \leq \alpha$. Thus, when the scores are allowed to reach values of order $\approx 1$, and when $m\leq n$, \Theorem{general_case} implies that the Lipschitz bound is also of order $\approx 1$. The assumptions (1)-(3) of \Theorem{general_case} are rather restrictive, and finding a proper normalization $c(X)$ in the general case is a difficult problem. However, we will see in the next section that a solution can be found in most practical cases of interest.

\section{The \emph{\MN{}} normalization}\label{sec:lipnorm}
\begin{figure}[t]
    \centering
    \includegraphics[width=0.48\textwidth]{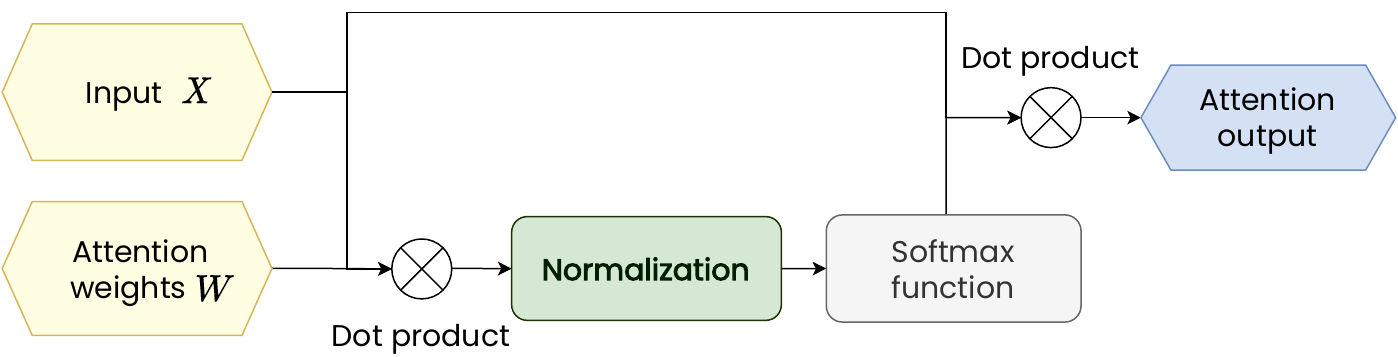}
    \caption{Pipeline of an attention mechanism along with the proposed normalization. For clarity, we assume a linear score function $g(x) = W^\top X$, expressed as a dot product operator.}
    \label{fig:norm_pipeline}
\end{figure}

In this section, we present our proposed normalization, the \MN{} in three different settings: Lipschitz, linear and quadratic score functions. These settings cover most of the practical applications, including Transformers, GAT and GT models. All settings are particular instances of a common idea: impose assumptions (1)-(3) of \Theorem{general_case} by dividing by the maximum of input values. A visualization of how \MN{} is applied to an attention model with linear score function is shown in \Fig{norm_pipeline}.

\paragraph{Lipschitz scores:}
When the score function is Lipschitz, the assumptions in \Theorem{general_case} can be met, for $\alpha\geq 0$, by 
\BEQ\label{eq:lip_score}
g(X) = \frac{\alpha\,\tilde{g}(X)}{\max\left\{\| \tilde{g}(X)\|_{(2,\infty)}, \|X^\top\|_{(\infty,2)}L_{F,(2,\infty)}(\tilde{g})\right\}}\,.
\EEQ
The denominator is composed of two terms: the first ensures that all the scores are bounded (by $\alpha$), while the second ensures that the gradient of the normalized scores remains low compared to the scale of the input vector. Note that $\|X^\top\|_{(\infty, 2)}$ is the maximum of the norm of input vectors.

\begin{theorem}\label{th:lip_score_th}
If the score function $\tilde{g}$ is Lipschitz continuous, then the attention layer with score function as defined in \Eq{lip_score} is Lipschitz continuous and
\BEQ
L_F(\Att) \leq e^{\alpha}\sqrt{\frac{m}{n}} + \alpha\sqrt{8}\,.
\EEQ

\end{theorem}

Note that $\alpha = 0$ leads to a uniform distribution and gradient vanishing when $m\ll n$ (for example $m=1$ when the output is a single vector). On the contrary, a large $\alpha \gg 1$ will lead to very large gradients that may destabilize training. In our experiments, we show that $\alpha=1$ is a good trade-off that allows to create relatively peaked attention weights, while maintaining a low Lipschitz constant (see \Sec{exps}).

\begin{figure*}[t]
    \centering
    \includegraphics[width=0.84\textwidth, height=0.38\textwidth]{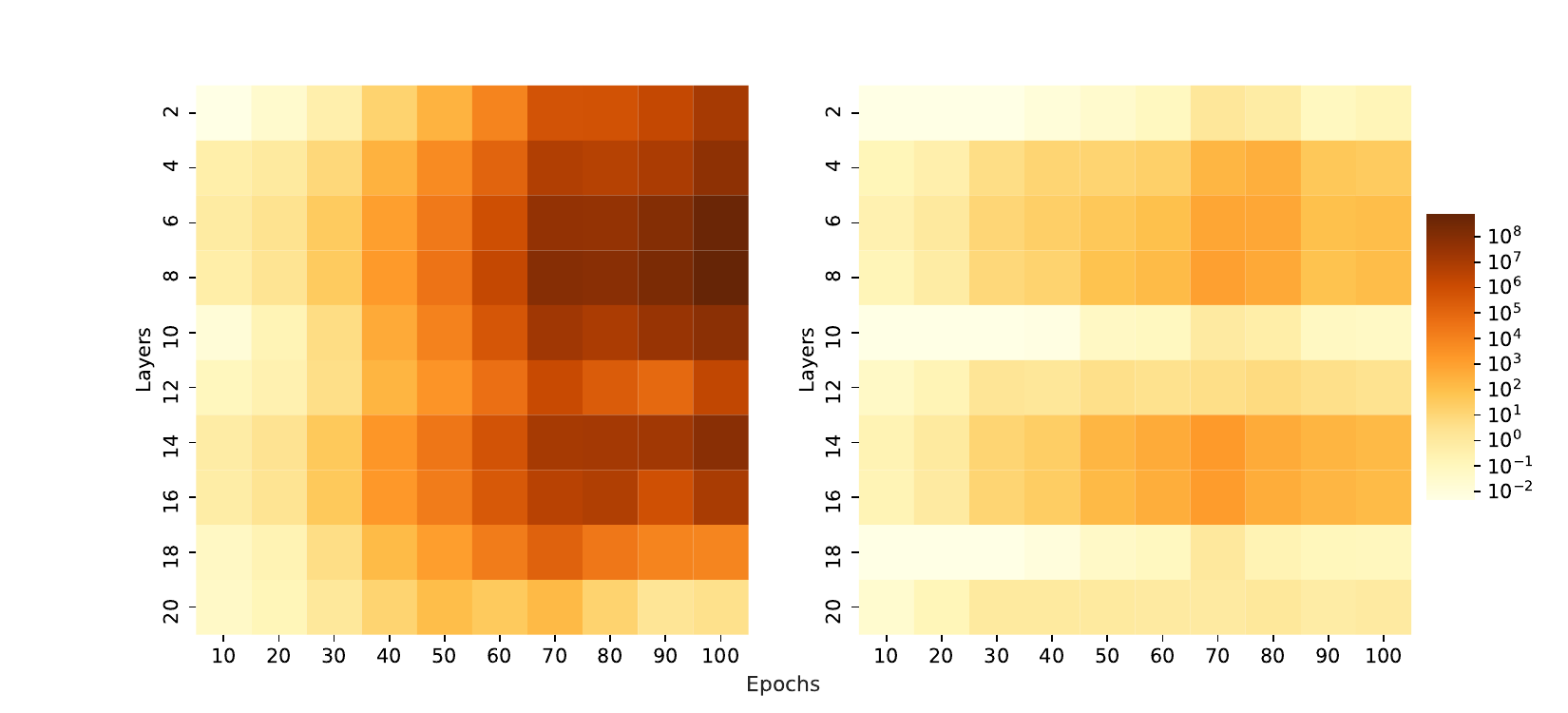}
    \caption{Gradient evolution of attention weights of a 20-layer GAT model for each layer throughout training. Each cell $i,j$ represents the norm of the gradients of the attention weights in the $i$-th layer and trained until $j$-th epoch. The left heat map corresponds to the standard GAT without any normalization, where the phenomenon of \textbf{gradient explosion} occurs. The right heat map corresponds to the GAT model using \MN{}. The proposed normalization restrains the attention weights from explosion.}
    \label{fig:grads_attention}
\end{figure*}

\paragraph{Linear Score Function:}
The initial definition of attention layers considers a linear score function $\tilde{g}(X) = Q^\top X$ for $Q\in\R^{d\times m}$. As this score function is Lipschitz continuous, \Theorem{lip_score_th} is directly applicable and leads to the following normalization, called \MN{},
\BEQ\label{eq:linear_score}
g(X) = \frac{Q^\top X}{\|Q\|_F\|X^\top\|_{(\infty,2)}}\,.
\EEQ
Note that, contrary to Transformers, the query matrix $Q$ is assumed to be a parameter of the model instead of an input.
\begin{corollary}
The attention layer with score function as defined in \Eq{linear_score} is Lipschitz continuous and
\BEQ
L_F(\Att) \leq e^{1}\sqrt{\frac{m}{n}} + \sqrt{8}\,.
\EEQ

\end{corollary}


\paragraph{Transformer Case:}
For Transformers, the query matrix $Q$ is an input of the model, and the score function is thus quadratic. As quadratic functions are not Lipschitz, \Theorem{lip_score_th} is not applicable. Fortunately, we can adapt the same idea to this setting.
As defined in \Sec{attentiondef}, let $X=(Q||K||V)$ be a concatenation of queries, keys and values.
Then, the assumptions in \Theorem{general_case} are met by
\BEQ\label{eq:quad_score}
g(X) = \frac{Q^\top K}{\max\left\{uv, uw, vw\right\}}\,,
\EEQ
where $u=\|Q\|_F$, $v=\|K^\top\|_{(\infty, 2)}$, and $w=\|V^\top\|_{(\infty, 2)}$.
Compared to the linear case of \Eq{linear_score}, we decompose the input matrix norm $\|X^\top\|_{(\infty, 2)}$ into $\|K^\top\|_{(\infty, 2)}$ and $\|V^\top\|_{(\infty, 2)}$ and return the product between the maximum and second maximum of $\|Q\|_F$, $\|K^\top\|_{(\infty, 2)}$, and $\|V^\top\|_{(\infty, 2)}$.

\begin{corollary}
The attention layer with score function as defined in \Eq{quad_score} is Lipschitz continuous and
\BEQ
L_F(\Att) \leq e^{\sqrt{3}}\sqrt{\frac{m}{n}} + 2\sqrt{6}\,.
\EEQ

\end{corollary}

Finally, as discussed in \Sec{attentiondef}, we normalize multi-head attention by normalizing each attention head separately. \Theorem{multiheadatt} then directly implies the following bound on the Lipschitz constant of the whole multi-head attention layer:
\BEQ
L_F(\MultAtt) \leq 11\|W_O\|_* \sqrt{\sum_{k=1}^h \|W_k\|_*^2}\,,
\EEQ
where $W_1,\dots,W_h$ (resp. $W_O$) represent the input (resp. output) projection matrices (see \Eq{multiheadatt}), and $m=n$.

\paragraph{Implementation details:}
Given an attention model $\mathcal{M}$ with score function $g$, we define $\mathcal{M}$-Lip as the updated attention model with the application of \MN{}. This normalization requires three steps, that we now provide for the linear and Transformer settings:
\begin{enumerate}
    \itemsep0em 
    \item \textbf{Frobenius norm of the queries}: First, we compute the Frobenius norm of $Q$: $u = \sqrt{\sum_i \|q_i\|_2^2}$.
    \item \textbf{Input norms}: Then, we compute the maximum 2-norm of the input vectors $v=\max_i \|x_i\|_2$ (or $v=\max_i \|k_i\|_2$ and $w=\max_i \|v_i\|_2$ for Transformers).
    \item \textbf{Scaling}: Finally, we divide the score function by the product $uv$ (or $\max\{uv, uw, vw\}$ for Transformers).
\end{enumerate}

Each attention head is treated separately, and thus all the norms and maximums are taken \emph{per head}. Moreover, in the case of graph attention, the norms and maximums are computed \textit{neighbor-wise}, i.e. for each node, we compute the maximum of the 2-norms of its neighbors. 

\paragraph{Complexity:} 
 The overcost of the proposed method is based on the row-wise and column-wise norm computations. Given that $n$ is the number of input vectors, $d$ is the representation dimensionality and $h$ the number of heads, the complexity of \MN   $\mathcal{O}(hnd)$. It remains negligible w.r.t. the overall cost of attention that is $\mathcal{O}(hnd^2)$.

\section{Gradient Explosion and Vanishing}
\label{sec:explosion}

Similar to the deep neural networks, the design and efficient training of deep attention models has a tight connection with their Lipschitz continuity. In fact, given $M$ Lipschitz continuous attention layers $\Att_m(\cdot)$ with Lipschitz constants $l_m = L_F(\Att_m)$ , their composition $f = \Att_1 \circ \Att_{2} \circ ... \circ \Att_{m-1} \circ \Att_{m}$  is Lipschitz continuous with Lipschitz constant upper-bounded by \BEQ\label{eq:composition_lip} L_F(f) \leq \prod_{m=1}^M L_F(\Att_m).\EEQ \Eq{composition_lip} implies that there is a multiplicative effect on the gradient flow of an M-layered attention model. Thus, enforcing the attention layer to be Lipschitz continuous  with tight Lipschitz bounds can alleviate gradient explosion and allow for the design of deeper attention-based models. 

\begin{figure}[t]
    \centering
    \includegraphics[width=0.5\textwidth]{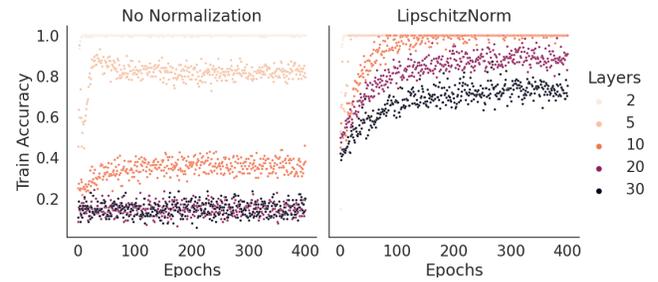}
    \caption{Convergence of train accuracy for a GAT model on node classification task using no normalization (left) and using \MN (right).}
    \label{fig:convergence}
\end{figure}

\Fig{grads_attention} (left picture) shows that gradient explosion occurs in a deep (20 layers) Graph Attention Network (GAT) \citep{gat} applied to a node classification task on Cora dataset~\citep{cora}. Throughout 100 epochs of training, the gradients of the attention weights in each GAT layer exhibit a steep increase, reaching extremely large value of the order of $10^8$.
However, \Fig{grads_attention} (right picture) shows that \MN{} is able to prevent gradient explosion, and throughout training, the gradients of the attention weights remain stable.
\Fig{convergence} shows that gradient explosion also comes with poor performance, and even a total lack of improvement in training accuracy for a GAT model with $30$ layers. Again, \MN{} avoids this behavior and allows for proper training in all regimes, 
showing that enforcing the Lipschitz continuity of the attention layer can help towards the design of deeper architectures.

\begin{table*}[t]
\centering
\caption{Classification accuracies for the missing-vector setting. In parentheses we denote the number of layers of the best model chosen for the highlighted accuracy. We denote by '-Pn' the application of PairNorm and by '-Lip' the application of the proposed \MN{}.}

\begin{tabular}{cccccccc}
\hline
                   & \multicolumn{2}{c}{\textbf{Cora}}       & \multicolumn{2}{c}{\textbf{CiteSeer}}   & \multicolumn{2}{c}{\textbf{Pubmed}}    \\ 
                   & \textbf{0\%}    & \textbf{100\%}      & \textbf{0\%}    & \textbf{100\%}      & \textbf{0\%}    & \textbf{100\%}     \\ \hline
\textbf{GCN}       & 82.5$\pm$ 1.2 (2) & 58.8 $\pm$ 3.5 (2)   & \textbf{69.5 $\pm$ 2.1 (2)}& 31.3 $\pm$ 2.7 (2) & 77.9 $\pm$ 1.4 (2)          & 44.9 $\pm$ 4.4 (2)          \\
\textbf{GGNN}       & 81.8 $\pm$ 2.0 (2)          & 68.2  $\pm$ 2.5 (6) & 68.5 $\pm$ 1.9 (3) & 40.5 $\pm$ 1.4 (5)           & 78.4 $\pm$ 2.1 (4)          & 56.6 $\pm$ 1.9 (4)          \\ 
\textbf{GAT}       & 82.3 $\pm$ 2.3 (2)          & 65.3 $\pm$ 2.1 (4)           & 69.3 $\pm$ 1.6 (2)          & 42.8 $\pm$ 1.6 (4)           & 77.4 $\pm$ 0.5 (6)  &        63.1 $\pm$ 0.7 (4)          \\ 
\textbf{GAT-Pn}   & 78.8 $\pm$ 0.6 (4) & 73.8 $\pm$ 1.2 (12)          & 67.2 $\pm$ 0.8 (4) &\textbf{51.7 $\pm$ 1.1 (10)} & 77.6 $\pm$ 1.6 (8) & 70.4 $\pm$ 1.1 (12)         \\
\textbf{GAT-Lip}   & \textbf{83.1$\pm$ 0.5 (5)}           & \textbf{75.3$\pm$ 0.9 (11)}  & 69.1 $\pm$ 1.5 (3)          & 50.9 $\pm$ 1.9 (9) & \textbf{78.9 $\pm$ 1.3 (5)} & \textbf{73.3 $\pm$ 1.4  (15)} \\ \hline
\end{tabular}
\label{tab:node_class_results}
\end{table*}

\section{Experimental Evaluation}\label{sec:exps}

We now examine the practical contribution of our normalization \MN{} in real-world and synthetic benchmarks.  In \Sec{exp_miss}, we evaluate \MN{} in real-world datasets that require the design of deeper GNN models. In \Sec{exp_trees}, we perform a synthetic study of increasing data and model depth and in \Sec{exp_realworld}, we apply \MN{} to attention-based GNNs of increasing model depth in real-world node classification tasks.

\subsection{Node Classification with Missing Information} \label{sec:exp_miss}
In most standard node classification benchmarks, the nodes present short-range dependencies, thus, making the fair evaluation of deeper models a difficult task. Towards a more solid comparison of deep GNNs,  \citet{pairnorm} presented a realistic framework that requires the design of deeper models by introducing an information noise of missing feature vectors. In particular, for a node classification task let a node attributed graph $\mathcal{D} = (V_u \cup V_l,E,X,U)$, where $V_u,V_l$ are the node sets of the unlabeled and the labeled nodes respectively, $E$ is the edge set, $X\in \mathbb{R} ^{n\times d}$ is the node attribute matrix and $U\in \mathbb{N}^{n \times m}$ is the label matrix. For an unlabeled node subset $\mathcal{M} \subseteq V_u$ we remove its node attributes: $\{X_j | j \in \mathcal{M}\}$ and we call this framework \textit{missing-vector setting} with fraction $p =\frac{|\mathcal{M}|}{|V_u|}$. This setting can represent cases of graph-based classification tasks with the cold-start phenomenon (i.e there is no history/feature information of the entities/nodes). 
 
\paragraph{Dataset and Model Setup:} We used three standard node classification datasets Cora, CiteSeer and PubMed~\citep{cora,citeseer}. The train/validation/test splits were the same as in~\cite{kipf}. Following~\citet{pairnorm}, for each dataset we had 2 node feature setups: the $0\%$ setup, where no attribute were removed and the $100\%$ setup, where all attributes of the unlabeled nodes were removed. We experimented with 3 models: 1) \textbf{GCN}: Graph Convolutional Network~\citep{kipf}, 2) \textbf{GGNN}: Gated Graph Neural Network~\citep{li2016gated} and 3) \textbf{GAT}: Graph Attention Network~\citep{gat}. A full description of the experimentation details is provided in the supplementary material.

\paragraph{Results:} Table \ref{tab:node_class_results} shows the average classification accuracy achieved in the standard and the missing-vector setting. \MN{} enables the training of deeper GAT layers, as in the missing-vector setting with $p = 100\%$, GAT-Lip (i.e. the GAT model with \MN{}) achieves state-of-the-art classification accuracies in four out of the six setups. Moreover, it is noteworthy that GAT-Lip exhibits a solid performance for both the $0\%$ and the $100\%$ scenarios, outperforming PairNorm.

\begin{figure}[t]
    \centering
    \includegraphics[width=.5\textwidth]{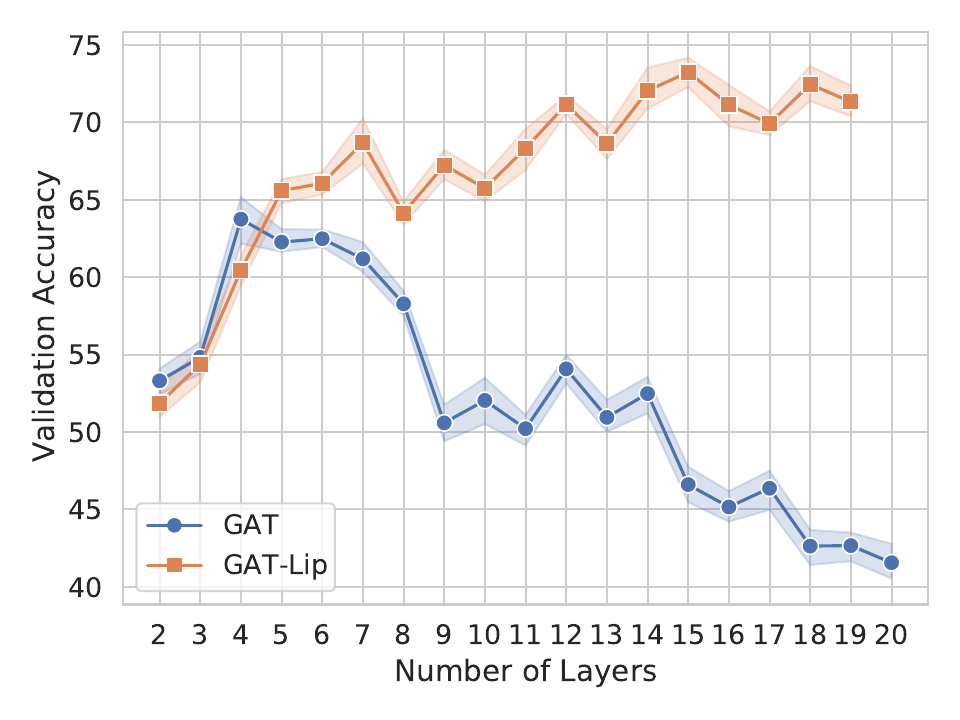}
    \caption{Classification accuracy of Graph Attention Network (GAT) with and without \MN for the $100\%$ setting of PubMed.}
    \label{fig:gat_pubmed_100}
\end{figure}

The need for a deeper GNN model is clear in Fig.~\ref{fig:gat_pubmed_100}. We visualize the performance of a GAT model with and without \MN in the $100\%$ setting of the PubMed dataset. Specifically, the GAT-Lip model exhibits an increasing accuracy as the number of layers is higher, showing that a larger depth is required for the inference in the case of the missing feature information. Also, it is clear that \MN has a crucial impact on the model training, as GAT without \MN fails to learn the task.
\paragraph{Deep Attention vs Deep Convolution:} \MN is a normalization that can be included in any attention model to establish Lipschitz bounds and build deeper architectures. It is interesting to see how a deep attention-based graph model is compared to a deep convolution-based one. Thus, we compare the GAT-Lip model with the GCNII~\citep{gcnii} in Cora, PubMed ($100\%$ setting) and in Ogbn-arxiv ($0\%$ setting) dataset~\citep{ogb}. 

\begin{table}[h!]\label{tab:gcnii_gat}
\centering
\caption{Comparison of GAT-Lip, GCNII for Ogbn-arxiv in $0\%$ setting and Cora/PubMed in $100\%$. Asterisk denotes the reported result in the public OGB leaderboard.}
\resizebox{0.49\textwidth}{!}{%

\begin{tabular}{cccc}
\hline
         & \textbf{Ogbn-arxiv ($0\%$)} & \textbf{Cora ($100\%$)}                & \textbf{PubMed ($100\%$)}              \\ \hline
GCNII  & 72.74 (-)$^*$ & 74.9 $\pm$ 0.4 (14) & \textbf{73.9 $\pm$ 0.3} (16) \\ 
GAT-Lip & \textbf{74.62 $\pm$ 1.1 (8)} & \textbf{75.3$\pm$ 0.9 (11)}  & 73.3 $\pm$ 1.4 (15)  \\ 
\hline

\end{tabular}
}
\end{table}

Table~\ref{tab:gcnii_gat} shows that GAT-Lip can outperform GCNII in the $0\%$ setting of Ogbn-arxiv with a margin of $>1.5\%$. Furthermore, on the $100\%$ settings of Cora and PubMed GAT-Lip and GCNII using a number of layers $> 10$ achieve similar accuracies without a clear lead.

\subsection{Model Depth in Synthetic Trees}\label{sec:exp_trees}

An intuitive way to show the ability of a deeper GNN model to capture long-range interactions is to generate synthetic graphs with nodes that are distant and have the same behavior. Thus, following~\citet{alon2020}, we create the TREES dataset. That is a set of directed trees (from the root to the leaves) of labeled nodes with increasing depth $d \in \{2,..., 10\}$, where the leaves of the tree are colored \textit{blue}, the root of the tree and the predecessors of the leaves are colored \textit{green} and the rest of the nodes remain uncolored. The task is to predict the label of the tree's root green node, according to the label of the other green nodes. In other words, the label of the root node is affected by the information from the leaves.

\begin{figure}[t]
    \centering
    \includegraphics[width=0.49\textwidth, height= 0.4\textwidth]{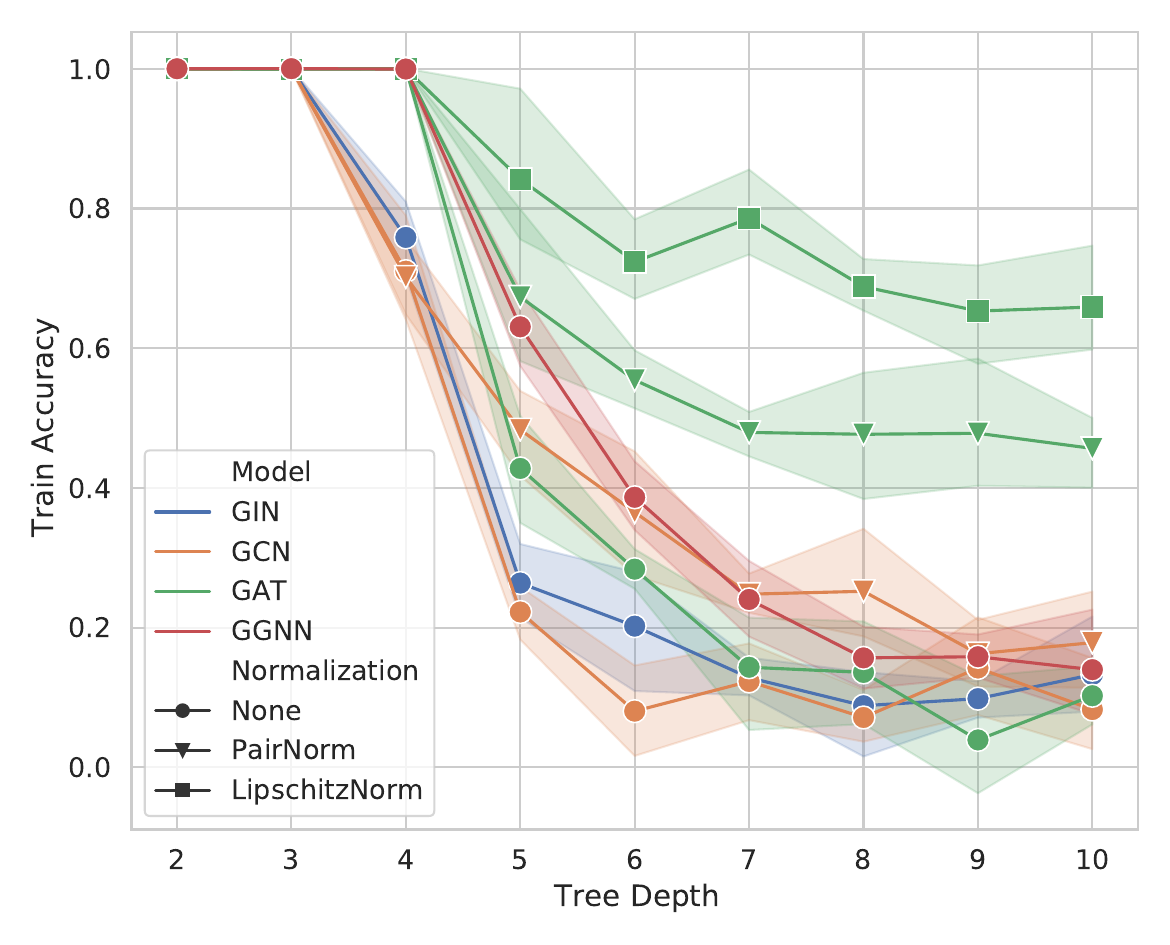}
    \caption{Train accuracies of four GNN models on the TREES dataset. We compare the model performance using: no normalization (circular dots), \textit{PairNorm} (triangular dots) and the proposed \textit{\MN{}} (square dots).}
    \label{fig:trees}
\end{figure}

\paragraph{TREES Dataset:} The generated tree structure simulate the exponential growth of the receptive field of the nodes, so that the information passes between two distant nodes. For this goal, we created for every tree depth 5000 binary trees and we run each experiment 10 times. Following \citet{alon2020}, we did not use explicitly extra blue neighbors, but, instead, we encoded their existence with 1-hot vectors of their cardinality as node attributes of the green nodes. 
\paragraph{Model Setup:} We compared the performance of Graph Convolutional Network~\citep{kipf}, Graph Isomorphism Network~\citep{xupowerful}, Gated Graph Neural Network~\citep{li2016gated} and Graph Attention Network~\citep{gat}. Each model was implemented with $d+1$ graph layers, where $d = \text{tree depth}$ and the hidden units size is set to 32. Moreover, we used either no normalization (None case in \Fig{trees}), PairNorm~\citep{pairnorm}, or our proposed \MN{}.

\paragraph{Discussion:}\Fig{trees} shows the train accuracy of the GNN models as the tree depth increases. Aligned with the previously found results, GAT and GGNN exhibit a better behavior with respect to the model depth. Moreover, the application of normalization methods has a significant impact on the performance of deeper models. GAT using the proposed normalization clearly outperforms the other architectures (for tree depth=10, GAT-Lip achieves $68.3\%$ training accuracy, while GAT-Pn achieves $47.8\%$ and all other variants achieve $<25\%$), showcasing the contribution of \MN{} to the design of deeper architectures.

\subsection{Model Depth in Real-World Datasets}\label{sec:exp_realworld}

In this section, we measure the behavior of \MN{} with respect to an increasing number of layers in real-world datasets. We apply the proposed normalization to two types of attention-based GNNs. We use the well-examined datasets \textit{Cora} and \textit{PubMed}~\citep{cora}  and two datasets from Open Graph Benchmark~\citep{ogb}: \textit{Ogbn-arxiv} and \textit{Ogbn-proteins}. Details and statistics of the datasets are provided in the supplementary material.

\begin{figure*}[t]
    \centering
    \includegraphics[width=1\textwidth]{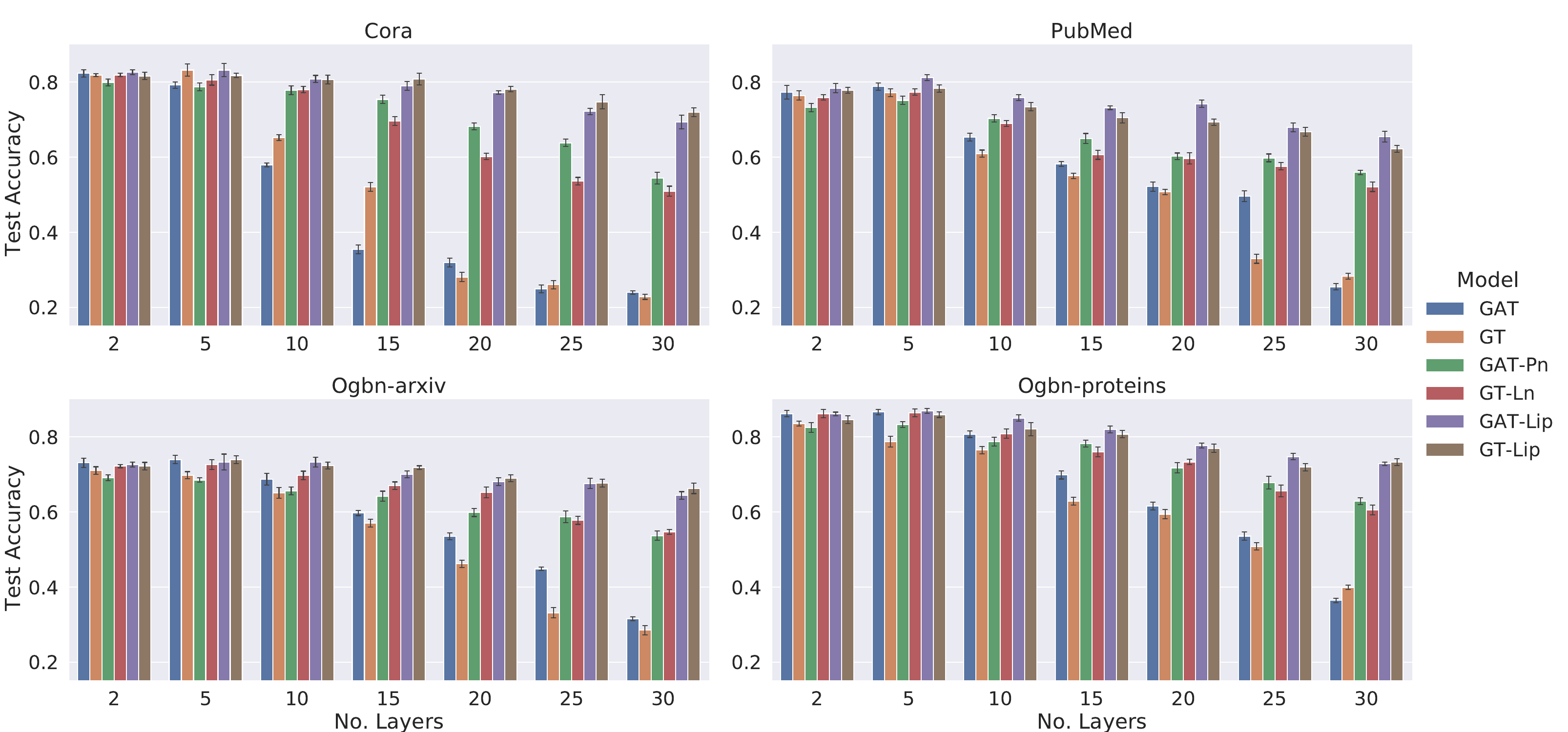}
    \caption{Test accuracies of a Graph Attention Network (GAT) and a Graph Transformer (GT). By '-Lip' we denote the application of \textit{\MN{}}, by '-Ln' the \textit{LayerNorm} and by '-Pn' the \textit{PairNorm}. In Ogbn-proteins dataset, the observed metric is ROC-AUC instead.}
    \label{fig:real}
\end{figure*}
\paragraph{Experimentation Setup:} We used two attention-based graph neural networks, as described in \Sec{lipnorm}: Graph Attention Network~\citep{gat} and Graph Transformer~\citep{unimp}. For the two models, we compared the contribution of \MN{} with two other normalization methods: the PairNorm~\citep{pairnorm} and the LayerNorm~\citep{layernorm}. The number of attention layers was $l \in \{2,5,10,15,20,25,30\}$. We performed cross validation, where the train/validation/test splits in Cora and PubMed were the same as in ~\citet{kipf}, and for Ogbn-arxiv and Ogbn-proteins we used the same splitting methods as used in~\citet{ogb}. The full experimental setup is described in the supplementary material.

\paragraph{Discussion:}In \Fig{real}, we highlight the impact of \MN{} on graph neural networks with respect to the model depth. For all four datasets \MN{} enables both GAT and GT to learn and maintain information throughout layers. Even for a large number of layers $(l >15)$, where the models without any normalization fail to converge, the variants using the \MN{} achieve comparable to the state-of-the-art results in the node classification tasks. More importantly, \MN{} outperforms PairNorm and LayerNorm, as it enhances the performance of shallow architectures and maintains it to deeper architectures.

\paragraph{Residual Connections:} Residual connections have been proven to be useful towards the design of deeper GNN models~\cite{deepgcns,gcnii}. Therefore, a comparison with \MN and an evaluation how they can be combined is necessary. Table~\ref{tab:residuals} suggests that residual connections with GAT layers enhance the performance of deep GNNs. However, LipschitzNorm significantly outperforms GAT-res in deep scenarios ($+6\%$ improvement on Cora with $30$ layers). Moreover, combining LipschitzNorm with residual connections slightly improves the performance, showing the ability of our method to be smoothly incorporated in various models.

\begin{table}[t]
\centering
\caption{Classification accuracies of deep GAT models (15 and 30 layers) with/without residual connections and with/without \MN.}
\label{tab:residuals}

\resizebox{0.48\textwidth}{!}{%
\tiny
\begin{tabular}{ccccc}
\hline
            & \multicolumn{2}{c}{\textbf{Cora}}                                           & \multicolumn{2}{c}{\textbf{PubMed}}                                        \\ 
Num. of layers              & \textbf{15}             & \textbf{30}                     & \textbf{15}            & \textbf{30}             \\
            \hline
GAT               & 38.2 $\pm$ 1.7          & 29.5$\pm$ 3.6       & 68.9$\pm$ 1.5          & 28.2 $\pm$ 3.1          \\
GAT-res     & 76.1 $\pm$ 1.2          & 63.5 $\pm$ 2.1      & 76.2 $\pm$ 1.1         & 63.8 $\pm$ 3.3          \\
GAT-Lip     & 79.4 $\pm$ 0.7          & \textbf{69.3 $\pm$ 4.1} & \textbf{76.4$\pm$ 1.5} & 67.2 $\pm$ 2.1          \\
GAT-Lip-res        & \textbf{80.2 $\pm$ 1.1} & \textbf{69.4 $\pm$ 2.8}& \textbf{77.3 $\pm$ 1.0}         & \textbf{68.7 $\pm$ 1.8} \\ \hline
\end{tabular}}
\end{table}

\section{Conclusion}

In this work, we introduced a novel normalization layer for attention-based neural networks, called \MN{}. We proved that the application of \MN{} enforces the Lipschitz continuity of self-attention layers.
In an empirical study, we showed that Lipschitz continuous modules can prevent from gradient explosion phenomena and, thus, can improve the performance of deep attention models.
Focusing on Graph Neural Networks (GNNs), where designing deep models is still a challenging task, we applied \MN{} to standard attention-based GNNs. We showed that \MN{} allows to build deep GNN architectures with strong performance on node classification tasks that exhibit long-range interactions.


\bibliography{biblio}
\bibliographystyle{icml2021}

\onecolumn
\begin{center}
\Large\bf
Supplementary material
\end{center}

\section{Proofs}

\subsection{Proof of Theorem 1}
\begin{theorem}\label{th:multiheadatt_supp}
If each attention head is Lipschitz continuous, then multi-head attention as defined in Eq.~(5) of the paper is Lipschitz continuous and
\BEQ
L_F(\MultAtt) \leq L_F(\Att) \|W_O\|_* \sqrt{\sum_{k=1}^h \|W_k\|_*^2}\,.
\EEQ
\end{theorem}
\begin{proof}
As $\MultAtt(X) = W_O \bigg(\Att(W_1 X)\bigg|\bigg|\dots\bigg|\bigg|\Att(W_h X)\bigg)$, we have, for any matrix $H\in\RR^{d\times n}$,
\BEQ
\jac\MultAtt_X(H) = W_O \bigg(\jac\Att_{W_1 X}(W_1 H)\bigg|\bigg|\dots\bigg|\bigg|\jac\Att_{W_h X}(W_h H)\bigg)\,.
\EEQ
First, note that, for any matrices $A\in\RR^{n\times m}$ and $B\in\RR^{m\times l}$, we have $\|AB\|_F^2=\sum_i \|A B_i\|_2^2 \leq \sum_i \|A\|_*^2\|B_i\|_2^2 = \|A\|_*^2\|B\|_F^2$ by definition of the spectral norm $\|A\|_*$. Hence, we have, for any matrices $X,H\in\RR^{d\times n}$,
\begin{align*}
\|\jac\MultAtt_X(H)\|_F^2 &\leq \|W_O\|_*^2 \left\|\bigg(\jac\Att_{W_1 X}(W_1 H)\bigg|\bigg|\dots\bigg|\bigg|\jac\Att_{W_h X}(W_h H)\bigg)\right\|_F^2\\
&= \|W_O\|_*^2 \sum_{k=1}^h \|\jac\Att_{W_k X}(W_k H)\|_F^2\\
&\leq \|W_O\|_*^2 \sum_{k=1}^h \vvvert\jac\Att_{W_k X}\vvvert_F^2\|W_k H\|_F^2\\
&\leq \|W_O\|_*^2 \sum_{k=1}^h L_F(\Att)^2\|W_k\|_*^2 \|H\|_F^2\,,
\end{align*}
which leads to the desired result, as $L_F(f) = \max_X \vvvert\jac f_X\vvvert_F = \max_{X,H} \|\jac f_X(H)\|_F / \|H\|_F$.
\end{proof}

\subsection{Proof of Lemma 1}
\begin{lemma}\label{lemma:derivative_bound_supp}
For any $X\in\R^{d\times n}$, the norm of the derivative of attention models (see Eq.~(4) of the paper) is upper bounded by:
\begin{equation}\label{eq:lf_bound1_supp}
\vvvert\jac \Att_X\vvvert_F \leq \|\softmax(g(X))\|_F + \sqrt{2}\|X^\top\|_{(\infty,2)}\vvvert\jac g_X\vvvert_{F, (2,\infty)}\,.
\end{equation}
\end{lemma}
\begin{proof}
Using the chain rule on the derivative of $\Att(X) = X \softmax(g(X))^\top$, we immediately obtain, for any $H\in\RR^{d\times n}$,
\begin{align}
\label{eq:chainrule.lemma.simple_supp}
\jac \Att_X(H) = H \softmax(g(X))^\top + X \jac\softmax_{g(X)}(\jac g_X(H))^\top\,,
\end{align}
and thus
\BEQ
\|\jac \Att_X(H)\|_F \leq \|H \softmax(g(X))^\top\|_F + \|X \jac\softmax_{g(X)}(\jac g_X(H))^\top\|_F\,.
\EEQ

First, we have $\|H \softmax(g(X))^\top\|_F\leq \|H\|_F\|\softmax(g(X))\|_F$ by multiplicativity of the Frobenius norm.
The second term follows from the bound, for any matrices $A\in\RR^{d\times n}$ and $B\in\RR^{m\times n}$,
\BEQ\label{eq:dsoftmaxbound_supp}
\|A \jac_X\softmax(B)^\top\|_F \leq \sqrt{2} \| A^\top \|_{(\infty, 2)}\|B\|_{(2,\infty)}\,\,,
\EEQ
that we will prove below.
Assuming that \Eq{dsoftmaxbound_supp} holds, we have, for any $H\in\RR^{d\times n}$,
\BEQ
\|\jac \Att_X(H)\|_F \leq \left(\|\softmax(g(X))\|_F + \sqrt{2} \|X^\top \|_{(\infty, 2)}\vvvert\jac g_X\vvvert_{F,(2,\infty)}\right)\|H\|_F\,,
\EEQ
and the desired result.
\Eq{dsoftmaxbound_supp} is proven as follows: the derivative of the softmax is given by
\BEQ
\jac_X\softmax(B)_{ij} = \sum_k \softmax(X)_{ij} \softmax(X)_{ik} (B_{ij} - B_{ik})\,,
\EEQ
and thus
\begin{align*}
    (A\jac_X\softmax(B)^\top)_{ij} &= \sum_k A_{ik}\jac_X\softmax(B)_{jk}\\
    &= \sum_{k,l} A_{ik}\softmax(X)_{jk} \softmax(X)_{jl} (B_{jk} - B_{jl})\\
    &= \sum_k A_{ik}\softmax(X)_{jk} B_{jk} - \sum_l \left(\sum_k A_{ik}\softmax(X)_{jk}\right) \softmax(X)_{jl} B_{jl} \\
    &= \sum_k A_{ik}\softmax(X)_{jk} B_{jk} - \sum_l (A\softmax(B)^\top)_{ij}\softmax(X)_{jl} B_{jl}\\
    &= \sum_k \softmax(X)_{jk} B_{jk} \left(A_{ik} - (A\softmax(B)^\top)_{ij}\right)\,.
\end{align*}
Inserting this last equality within the Frobenius norm, we get
\begin{align*}
        \|A(\jac_X\softmax(B))^\top\|_F^2 &= \sum_{i,j} \left(\sum_k \softmax(X)_{jk} B_{jk} (A_{ik} - (A\softmax(B)^\top)_{ij})\right)^2 \\
        &\leq \sum_{i,j,k} \softmax(X)_{jk} B_{jk}^2 \left(A_{ik} - (A\softmax(B)^\top)_{ij}\right)^2 \\
    &= \sum_{j,k} \softmax(X)_{jk} B_{jk}^2 \|A^\top_k - (A\softmax(B)^\top)^\top_j\|_2^2\,,
\end{align*}
where the inequality comes from Jensen's inequality applied to the square function (i.e. $\mathbb{E}[Z]^2\leq\mathbb{E}[Z^2]$ for any r.v. $Z$) and the fact that $\softmax(X)_j$ is a probability distribution. Finally, as $(A\softmax(B)^\top)^\top_j$ is a weighted average of the vectors $A^\top_k$, and is thus in their convex hull, we have $\|A^\top_k - (A\softmax(B)^\top)^\top_j\|_2^2 \leq 2 \max_k \|A_k^\top\|_2^2$, and thus
\begin{align*}
\|A(\jac_X\softmax(B))^\top\|_F^2 &\leq 2\sum_{j,k} \softmax(X)_{jk} B_{jk}^2 \|A^\top\|_{(\infty,2)}^2\\
&\leq 2 \sum_j \|\softmax(X)^\top_j\|_1 \|B^\top_j\|_\infty^2 \|A^\top\|_{(\infty,2)}^2\\
&= 2 \|B\|_{(2,\infty)}^2 \|A^\top\|_{(\infty,2)}^2\,,
\end{align*}
where the second inequality uses the H\"older inequality and the last line is due to $\|\softmax(X)^\top_j\|_1 = 1$. This finishes the proof and leads to the desired inequality.
\end{proof}

\subsection{Proof of Lemma 2}
\begin{lemma}
\label{lemma:softmax.frobenius.bounds_supp}
For any $M \in \RR^{m \times n}$, we have
\BEQ
\sqrt{m/n} \leq \|\softmax(M)\|_F \leq \sqrt{m}\,.
\EEQ
\end{lemma}
\begin{proof}
Let $M \in \RR^{m \times n}$, we have
\begin{align*}
    \| \softmax(M) \|_F = \left( \sum_{i,j} \softmax(M)_{ij}^2 \right)^{1/2}\,.
\end{align*}
First, using $\softmax(M)_{ij}\in[0,1]$ and $\sum_j \softmax(M)_{ij} = 1$ for all $i$, we obtain
\BEQ
\|\softmax(M)\|_F = \sqrt{\sum_{i,j}  \softmax(M)_{ij}^2} \leq \sqrt{\sum_{i,j}  \softmax(M)_{ij}} = \sqrt{m}\,.
\EEQ
Then, using Eq.(10) of the paper, we have
\BEQ
\|\softmax(g(X))\|_F = \sqrt{\frac{m + \sum_{i=1}^m d_{\chi^2}(S_i, U_n)}{n}} \geq \sqrt{\frac{m}{n}}\,,
\EEQ
as the $\chi^2$-divergences are positive (i.e. $d_{\chi^2}(S_i, U_n)\geq 0$).
\end{proof}

\subsection{Proof of Lemma 3}
\begin{lemma}
If all the scores are bounded by $\alpha\geq 0$, i.e. for all $i\in\{1,\dots,m\}$ and $j\in\{1,\dots,n\}$, $|g(x)_{ij}|\leq \alpha$, then
\BEQ
\|\softmax(g(X))\|_F \leq e^{\alpha}\sqrt{\frac{m}{n}}\,.
\EEQ
\end{lemma}
\begin{proof}
If, for all $i\in\{1,\dots,m\}$ and $j\in\{1,\dots,n\}$, we have $|g(x)_{ij}|\leq\alpha$, then
\BEQ
\softmax(g(X))_{ij} = \frac{e^{g(X)_{ij}}}{\sum_k e^{g(X)_{ik}}} \leq \frac{e^{\alpha}}{e^{\alpha} + (n-1)e^{-\alpha}} = \frac{1}{1 + (n-1)e^{-2\alpha}} \leq \frac{e^{2\alpha}}{n}\,.
\EEQ
Hence, we have
\BEQ
\|\softmax(g(X))\|_F = \sqrt{\sum_{i,j} \softmax(g(X))_{ij}^2} \leq  \sqrt{\sum_{i,j} \softmax(g(X))_{ij} \frac{e^{2\alpha}}{n}} = e^{\alpha}\sqrt{\frac{m}{n}}\,.
\EEQ
\end{proof}

\subsection{Proof of Theorem 2}
\begin{theorem}\label{th:general_case_supp}
Let $\alpha\geq 0$. If, for all $X\in\RR^{d\times n}$, we have
\BNUM
\itemsep0em
\item[(1)] $\|\tilde{g}(X)\|_{\infty}\leq \alpha c(X)$,
\item[(2)] $\|X^\top\|_{(\infty,2)}\vvvert\jac \tilde{g}_X\vvvert_{F,(2,\infty)}\leq \alpha c(X)$,
\item[(3)] $\|X^\top\|_{(\infty,2)}\vvvert\jac c_X\vvvert_{F,1}\|\tilde{g}(X)\|_{(2,\infty)}\leq \alpha c(X)^2$,
\ENUM
then attention models (see Eq.~(4) of the paper) with score function $g(X) = \tilde{g}(X) / c(X)$ is Lipschitz continuous and
\BEQ
L_F(\Att) \leq e^{\alpha}\sqrt{\frac{m}{n}} +  \alpha\sqrt{8}\,.
\EEQ
\end{theorem}
\begin{proof}
Using Lemma 1 and Lemma 3 and the assumptions (1), we have,
\begin{align*}
\vvvert\jac \Att_X\vvvert_F &\leq \|\softmax(g(X))\|_F + \sqrt{2}\|X^\top\|_{(\infty,2)}\vvvert\jac g_X\vvvert_{F, (2,\infty)}\\
&\leq e^{\alpha}\sqrt{\frac{m}{n}} + \sqrt{2}\|X^\top\|_{(\infty,2)}\vvvert\jac g_X\vvvert_{F, (2,\infty)}\,,
\end{align*}
where the first inequality is due to Lemma 1 and the second inequality is due to Lemma 3 and assumption (1) (as then $\|g(X)\|_{\infty}\leq \alpha$). Moreover, the derivative of the score function $g(X)=\tilde{g}(X)/c(X)$ gives
\BEQ
\jac g_X(H) = \frac{\jac\tilde{g}_X(H)}{c(X)} - \frac{\jac c_X(H)\tilde{g}(X)}{c(X)^2}\,,
\EEQ
and thus,
\BEQ
\vvvert\jac g_X\vvvert_{F,(2,\infty)} \leq \frac{\vvvert\jac \tilde{g}_X\vvvert_{F,(2,\infty)}}{c(X)} + \frac{\vvvert\jac c_X\vvvert_{F,1}\|\tilde{g}(X)\|_{(2,\infty)}}{c(X)^2}\,.
\EEQ
Finally, using this equation and assumption (2) and (3), we have
\begin{align*}
\vvvert\jac \Att_X\vvvert_F &\leq e^{\alpha}\sqrt{\frac{m}{n}} + \frac{\sqrt{2}\|X^\top\|_{(\infty,2)}\vvvert\jac \tilde{g}_X\vvvert_{F,(2,\infty)}}{c(X)} + \frac{\sqrt{2}\|X^\top\|_{(\infty,2)}\vvvert\jac c_X\vvvert_{F,1}\|\tilde{g}(X)\|_{(2,\infty)}}{c(X)^2}\\
&\leq e^{\alpha}\sqrt{\frac{m}{n}} + \sqrt{2}\alpha + \sqrt{2}\alpha\\
&\leq e^{\alpha}\sqrt{\frac{m}{n}} + \alpha\sqrt{8}\,,
\end{align*}
and the desired result.
\end{proof}

\begin{remark}\label{rem:hcontractive}
Note that Theorem 2 still holds if $\Att(X)=h(X)\,\softmax(g(X))^\top$ and $L_F(h)\leq 1$ (i.e. the function $h$ is contractive). In such a case, the assumptions become:
\BNUM
\itemsep0em
\item[(1)] $\|\tilde{g}(X)\|_{\infty}\leq \alpha c(X)$,
\item[(2)] $\|h(X)^\top\|_{(\infty,2)}\vvvert\jac \tilde{g}_X\vvvert_{F,(2,\infty)}\leq \alpha c(X)$,
\item[(3)] $\|h(X)^\top\|_{(\infty,2)}\vvvert\jac c_X\vvvert_{F,1}\|\tilde{g}(X)\|_{(2,\infty)}\leq \alpha c(X)^2$.
\ENUM
\end{remark}

\subsection{Proof of Theorem 3}
\begin{theorem}\label{th:lip_score_th_supp}
If the score function $\tilde{g}$ is Lipschitz continuous, then the attention layer with score function as defined in Eq.~(15) of the paper is Lipschitz continuous and
\BEQ
L_F(\Att) \leq e^{\alpha}\sqrt{\frac{m}{n}} + \alpha\sqrt{8}\,.
\EEQ
\end{theorem}
\begin{proof}
First, as $c(X) = \max\left\{\|\tilde{g}(X)\|_{(2,\infty)}, \|X^\top\|_{(\infty,2)}L_{F,(2,\infty)}(\tilde{g})\right\} / \alpha$, we have $\alpha c(X) \geq \|\tilde{g}(X)\|_{(2,\infty)} \geq \|\tilde{g}(X)\|_\infty$ and assumption (1) of Theorem 2 is verified.
Second, we have $\alpha c(X) \geq \|X^\top\|_{(\infty,2)}L_{F,(2,\infty)}(\tilde{g}) \geq \|X^\top\|_{(\infty,2)}\vvvert\jac\tilde{g}_X\vvvert_{F,(2,\infty)}$ and assumption (2) of Theorem 2 is also verified.
Finally, we have
\begin{align*}
\alpha|\jac c_X(H)| &\leq \max\left\{\left|\jac {\|\tilde{g}(\cdot)\|_{(2,\infty)}}_X(H)\right|, \left|\jac {\|\cdot^\top\|_{(\infty, 2)}}_X(H)\right| L_{F,(2,\infty)}(\tilde{g})\right\}\\
&\leq \max\left\{\|\jac\tilde{g}_X(H)\|_{(2,\infty)}, \|H^\top\|_{(\infty, 2)} L_{F,(2,\infty)}(\tilde{g})\right\}\\
&\leq \max\left\{\vvvert\jac\tilde{g}_X\vvvert_{F,(2,\infty)}\|H\|_F, \|H\|_F L_{F,(2,\infty)}(\tilde{g})\right\}\\
&\leq L_{F,(2,\infty)}(\tilde{g}) \|H\|_F\,,
\end{align*}
where the second inequality follows from the triangle inequality $\left|\|X+H\|_{(\infty, 2)} - \|X\|_{(\infty, 2)}\right|\leq \|H\|_{(\infty, 2)}$, implying that $|\jac {\|\cdot\|_{(\infty, 2)}}_X(H)|\leq \|H\|_{(\infty, 2)}$. As a result, we have $\|X^\top\|_{(\infty,2)}\vvvert\jac c_X\vvvert_{F,1}\|\tilde{g}(X)\|_{(2,\infty)}\leq \|X^\top\|_{(\infty,2)}L_{F,(2,\infty)}(\tilde{g})\|\tilde{g}(X)\|_{(2,\infty)}/\alpha \leq \alpha c(X)^2$ (using assumption (1) and (2)) and assumption (3) of Theorem 2 is also verified. We can thus apply Theorem 2 and obtain the desired result.
\end{proof}

\subsection{Proof of Corollary 1}
\begin{corollary}
The attention layer with score function as defined in Eq.~(17) of the paper is Lipschitz continuous and
\BEQ
L_F(\Att) \leq e^{1}\sqrt{\frac{m}{n}} + \sqrt{8}\,.
\EEQ
\end{corollary}
\begin{proof}
First, note that replacing $L_{F,(2,\infty)}(\tilde{g})$ in Theorem 2 by any upper bound $M \geq L_{F,(2,\infty)}(\tilde{g})$ does not change the result and, as $L_{F,(2,\infty)}(\tilde{g})=\|Q\|_*$ is hard to compute, we instead prefer the upper bound $\|Q\|_F \geq \|Q\|_*$ that is simple and fast to compute.
As $\tilde{g}(X)=Q^\top X$ is Lipschitz, we can directly apply Theorem 3 with $\alpha=1$ and $c(X) = \max\left\{\|Q^\top X\|_{(2,\infty)}, \|X^\top\|_{(\infty,2)}\|Q\|_F\right\}$ to get the desired result.
Moreover, the normalization simplifies to $c(X) = \|Q\|_F \|X^\top\|_{(\infty,2)}$, as $\|Q^\top X\|_{(2,\infty)} \leq \|Q\|_F \|X^\top\|_{(\infty,2)}$.
\end{proof}

\subsection{Proof of Corollary 2}
\begin{corollary}
The attention layer with score function as defined in Eq.~(19) of the paper is Lipschitz continuous and
\BEQ
L_F(\Att) \leq e^{\sqrt{3}}\sqrt{\frac{m}{n}} + 2\sqrt{6}\,.
\EEQ
\end{corollary}
\begin{proof}
As defined in Sec. 3.3 of the paper, let $X=(Q||K||V)$ be a concatenation of queries, keys and values, and $\Att(X) = V \softmax\left(g(X)\right)^\top$.
First, note that $\Att(X) = h(X)\,\softmax\left(g(X)\right)^\top$, where $h:X=(Q||K||V)\mapsto V$ is a projection. As projections are contractive, Remark~\ref{rem:hcontractive} implies that Theorem 2 can be used in such a case if we replace $\|X^\top\|_{(\infty,2)}$ by $\|V^\top\|_{(\infty,2)}$ in assumptions (1)-(3).
As proposed in Eq.(19) of the paper, let $g(X) = \tilde{g}(X) / c(X)$ where $\tilde{g}(X)=Q^\top K$, $c(X)=\max\left\{uv, uw, vw\right\}$, $u=\|Q\|_F$, $v=\|K^\top\|_{(\infty, 2)}$, and $w=\|V^\top\|_{(\infty, 2)}$. Then, we have
\begin{align*}
\|Q^\top K\|_\infty \leq \|Q^\top K\|_{(2,\infty)} \leq \|Q\|_F \|K^\top\|_{(\infty, 2)} = uv \leq c(X)\,,
\end{align*}
and assumption (1) is verified (with $\alpha=1$). Moreover, for any perturbation $H=(H_Q||H_K||H_V)$, where $H_Q$, $H_K$ and $H_V$ are the perturbations associated to, respectively, $Q$, $K$ and $V$, we have
\begin{align*}
\|D\tilde{g}_X(H)\|_{(2,\infty)} &\leq \|Q^\top H_K\|_{(2,\infty)} + \|H_Q^\top K\|_{(2,\infty)}\\
&\leq \|Q\|_F\|H_K^\top\|_{(\infty,2)} + \|H_Q\|_F\|K^\top\|_{(\infty,2)}\\
&\leq u\|H_K\|_F + v\|H_Q\|_F\\
&\leq \sqrt{u^2 + v^2} \|H\|_F\,,
\end{align*}
where the last inequality is due to the Cauchy-Schwarz inequality. Hence, we have
\begin{align*}
\|V^\top\|_{(\infty,2)}\vvvert\jac \tilde{g}_X\vvvert_{F,(2,\infty)} \leq w \sqrt{u^2 + v^2} \leq \sqrt{2}c(X)\,,
\end{align*}
and assumption (2) is verified (with $\alpha=\sqrt{2}$).
Finally, we have
\begin{align*}
|\jac c_X(H)| &\leq \max\{v,w\}\|H_Q\|_F + \max\{u,w\}\|H_K^\top\|_{(\infty, 2)} + \max\{u,v\}\|H_V^\top\|_{(\infty, 2)}\\
&\leq \max\{v,w\}\|H_Q\|_F + \max\{u,w\}\|H_K\|_F + \max\{u,v\}\|H_V\|_F\\
&\leq \sqrt{\max\{v,w\}^2 + \max\{u,w\}^2 + \max\{u,v\}^2} \|H\|_F\,,
\end{align*}
where the last inequality is due to the Cauchy-Schwarz inequality, and thus
\begin{align*}
\|\tilde{g}(X)\|_{(2,\infty)}\|V^\top\|_{(\infty,2)}\vvvert\jac c_X\vvvert_{F,1} \leq uvw \sqrt{\max\{v,w\}^2 + \max\{u,w\}^2 + \max\{u,v\}^2} \leq \sqrt{3} uvw\max\{u,v,w\} \leq \sqrt{3}c(X)^2\,,
\end{align*}
and assumption (3) is verified (with $\alpha=\sqrt{3}$). Hence, Theorem 2 with $\alpha=\sqrt{3}$ is applicable and immediately provides the desired result.
\end{proof}

\newpage
\section{Experiments}
In this section we report the dataset and experimentation setup in Section 7.1 and Section 7.3.

\subsection{Datasets}
Here, we present the details of the examined real-world datasets, that were used in Section 7.1 and Section 7.3.
\begin{itemize}
    \item \textbf{Cora, CiteSeer and PubMed} are citation networks~\cite{cora,citeseer}. Nodes correspond to research publications and edges encode citation links. All three datasets contain node attributes, that are sparse bag-of-words representations for each document (1433,3703, and 500-dimensional respectively).
    \item \textbf{Ogbn-proteins:} is a proteins interactions network, where each node representts a protein and each edge indicate biological interactions between proteins (e.g homology, co-expression, etc.)~\cite{ogb}. The dataset contains 8-dimensional edge attributes, where each dimension corresponds to the strength of the interaction type and 8-dimensional node attributes, that is one-hot encodings of the 8 species that a protein comes from. 
    \item \textbf{Ogbn-arxiv} is a citation network with directed edges, where each node corresponds to an arXiv paper and the edges denote citations from one paper to another~\cite{ogb}. The dataset contains node attributes, that are averaged word embeddings of the titles and the abstracts of dimensionality 128. The label of each node is the subject area of the paper and can take 40 values.
\end{itemize}

In Table~\ref{tab:dataset_statistics}, we report the statistics of the datasets. 
\begin{table}[h!] 
\centering
\caption{Datasets statistics. All datasets consist of a single graph. All node classification tasks are single-label, except \textit{Ogbn-proteins}, which is multi-label. Attributes correspond to node features, except for \textit{Ogbn-proteins} dataset, where attributes are a summation over the node and edge features.}
\label{tab:dataset_statistics}
\begin{tabular}{ccccc}
\hline
\textbf{Dataset}     & \textbf{\# Nodes} & \textbf{\# Edges}                     &  \textbf{\# Attributes}    & \textbf{\# Classes} \\ \hline
\textbf{Cora}                   & 2,708                       & 5,429    & 1433                                             & 7                   \\ 
\textbf{CiteSeer}            & 3,327                       & 4,732  & 3703                                               & 6                   \\ 
\textbf{PubMed}              &  19,717                     & 44,338   & 500                                              & 3                   \\ 

\textbf{Ogbn-arxiv}           & 169,343                     & 1,166,243 & 128 & 40                  \\ 
\textbf{Ogbn-proteins}           & 132,534                     & 39,561,252 & 16 & 2 (112-label)                  \\ \hline\end{tabular}

\end{table}

\subsection{Experimentation details for missing-vector setting}
Next, we present the experimentation setup that was followed in Section 7.1. This experiment corresponds to a \textbf{node classification task} under the \textit{missing vector} setting, as suggested in~\citealp{pairnorm}.
In our experiments, we used the Adam optimizer~\cite{adam} with a weight decay $L = 5*10^{-4}$ and the initial learning rate was set in $\{0.1,0.01,0.005,0.001\}$. We have run each experiment $5$ times for $1000$ epochs. The evaluation metric is the standard validation classification accuracy for all three datasets.
\paragraph{Model Selection:} For all three GNN models, i.e GCN, GGNN, GAT and the normalization scenarios we performed cross-validation with predefined train/validation/test splits. For a fair comparison we used the same splits for all three datasets (Cora, CiteSeer and PubMed) as reported and used in~\citealp{kipf}.
\paragraph{Hyper-parameter tuning:} We performed grid-search to tune the hyper parameters. The hyper-parameters that were tuned are the following:
\begin{itemize}
    \item \textbf{Number of GNN layers}: For all models and datasets, we used $l$ GNN layers where $l\in\{ 1,2,3,..,20\}$.
    \item \textbf{Hidden units size}: The dimensionality of the hidden units in all models was in $\{8,16,32,64,128\}$.
    \item \textbf{Attention heads}: In the case of the GAT model, the attention heads that we used were in $\{1,2,4,8\}$.
    \item \textbf{Dropout ratio}: The dropout ratio was set in $\{0, 0.5\}$.

\end{itemize}

\subsection{Experimentation details for real-world datasets with respect to the model depth}
In this section, we present the setup of the experimentation in Section 7.3. Same with Section 7.1, this experiment is a \textbf{node classification task}, where we evaluate the performance of GNN models with respect to increasing model depth. We used again the Adam optimizer~\cite{adam} with a weight decay $L = 5*10^{-4}$ and the initial learning rate was set in $\{0.1,0.01,0.005,0.001\}$.

\paragraph{Model Selection:} We performed, again, for all models and datasets cross-validation with predefined train/validation/test splits and reported the best achieved validation accuracy. For Cora and PubMed, as in Section 7.1, we used the same splits as in~\citealp{kipf}. For the other two datasets we have:
\begin{enumerate}
    \item \textit{Ogbn-arxiv}: We used the same splitting method as used in~\citealp{ogb}. Specifically, the train split corresponds to the papers published until 2017, the validation split to the ones published in 2018 and the test split to the ones published in 2019. We used a full-batch training.
    \item \textit{Ogbn-proteins} For this dataset, we used, also, the same splitting method as in~\citealp{ogb}. That is we split the nodes according to the node labels and in particular grouping according to the protein species. Similar to~\citealp{unimp}, we used neighbor sampling~\citep{graphsage} as a sampling method, due to the size of the graph.
\end{enumerate}

\paragraph{Model Depth:} In order to examine the model behavior under the depth increase, for each architecture we used models consisting of $l$ GNN layers, where $l \in\{2,5,10,15,20,25,30\}$. We run each experiment 5 times and we keep the configuration with the best average accuracy.

\paragraph{Hyper-parameter tuning:} For each model depth and GNN model, we performed grid-search for hyper-parameter tuning. The hyper-parameters that were tuned are the following:
\begin{enumerate}
    \item \textbf{Graph Attention Network}~\cite{gat}: The dimensionality of the hidden units was set in $\{8,16,64,128\}$. The number of attention heads was selected between $\{1,2,4,8\}$ and we experimented over two standard aggregators of the attention heads: a) \textit{concatenation} and b) \textit{averaging} of the attention heads. The dropout of the attention weights was set in $\{0,0.2,0.5\}$.
    \item \textbf{Graph Transformer} from the UNIMP framework~\cite{unimp}: The hidden dimensionality was selected from $\{8,16,64,128\}$ and the number of attentions heads from $\{1,2,4\}$. We tested \textit{concatenation} and \textit{averaging} of the attention heads and the dropout of the attention weights was set in $\{0,0.5\}$.
\end{enumerate}


\end{document}